# On the Convergence of Bound Optimization Algorithms

Ruslan Salakhutdinov
Sam Roweis
University of Toronto
6 King's College Rd, M5S 3G4, Canada
rsalakhu,roweis@cs.toronto.edu

Zoubin Ghahramani
Gatsby Computational Neuroscience Unit
University College London
17 Queen Square, London WC1N 3AR, UK
zoubin@gatsby.ucl.ac.uk

## Abstract

Many practitioners who use EM and related algorithms complain that they are sometimes slow. When does this happen, and what can be done about it? In this paper, we study the general class of *bound optimization* algorithms – including EM, Iterative Scaling, Non-negative Matrix Factorization, CCCP – and their relationship to direct optimization algorithms such as gradient-based methods for parameter learning. We derive a general relationship between the updates performed by bound optimization methods and those of gradient and second-order methods and identify analytic conditions under which bound optimization algorithms exhibit quasi-Newton behavior, and under which they possess poor, first-order convergence. Based on this analysis, we consider several specific algorithms, interpret and analyze their convergence properties and provide some recipes for preprocessing input to these algorithms to yield faster convergence behavior. We report empirical results supporting our analysis and showing that simple data preprocessing can result in dramatically improved performance of bound optimizers in practice.

## 1 Bound Optimization Algorithms

Many problems in machine learning and pattern recognition ultimately reduce to the optimization of a scalar valued function $L(\Theta)$ of a free parameter vector $\Theta$. For example, in supervised and unsupervised probabilistic modeling the objective function may be the (conditional) data likelihood or the posterior over parameters. In discriminative learning we may use a classification or regression score; in reinforcement learning an average discounted reward. Optimization may also arise during inference; for example we may want to reduce the cross entropy between two distributions or minimize a function such as the Bethe free energy.

Bound optimization (BO) algorithms take advantage of the fact that many objective functions arising in practice have a special structure. We can often exploit this structure to obtain a *bound* on the objective function and proceed by optimizing this bound. Ideally, we seek a bound that is valid everywhere in parameter space, easily optimized, and equal to the true objective function at one (or more) point(s).

A general form of a bound *maximizer* which iteratively *lower bounds* an objective function $L(\Theta)$ is given below:

---
**General Bound Optimizer for maximizing $L(\Theta)$:**
* **Assume:** $\exists\, G(\Theta, \Psi)$ such that for any $\Theta'$ and $\Psi'$:

  1. $G(\Theta', \Theta') = L(\Theta')$ & $L(\Theta) \geq G(\Theta, \Psi')\ \forall\, \Psi' \neq \Theta$
  2. $\arg\max_\Theta G(\Theta, \Psi')$ can be found easily for any $\Psi'$.

* **Iterate:** $\Theta^{t+1} = \arg\max_\Theta G(\Theta, \Theta^t)$
* **Guarantee:** $L(\Theta^{t+1}) = G(\Theta^{t+1}, \Theta^{t+1}) \geq$
  $G(\Theta^{t+1}, \Theta^t) \geq G(\Theta^t, \Theta^t) = L(\Theta^t)$
---

Bound optimizers do nothing more than coordinate ascent in the functional $G(\Theta, \Psi)$, alternating between maximizing $G$ with respect to $\Psi$ for fixed $\Theta$ and with respect to $\Theta$ for fixed $\Psi$. These algorithms enjoy a strong guarantee; they never worsen the objective function.

Many popular iterative algorithms are bound optimizers, including the EM algorithm for maximum likelihood learning in latent variable models[2], iterative scaling (IS) algorithms for parameter estimation in maximum entropy models[1], non-negative matrix factorization (NMF)[3] and the recent CCCP algorithm for minimizing the Bethe free energy in approximate inference problems[12].

In this paper we explore two questions of theoretical and practical interest: when will bound optimization be fast or slow relative to other standard approaches, and what can be done to improve convergence rates of these algorithms when they are slow?

## 2 Convergence Behavior and Analysis

How large are the steps that bound optimization methods take? Any bound optimizer implicitly defines a mapping: $M : \Theta \to \Theta'$ from parameter space to itself, so that $\Theta^{t+1} = M(\Theta^t)$. If iterates $\Theta^t$ converge to a fixed point $\Theta^*$, then $\Theta^* = M(\Theta^*)$. If $M(\Theta)$ is continuous and differentiable, we can Taylor expand it in the neighborhood of the fixed point $\Theta^*$:

$$\Theta^{t+1} - \Theta^* \approx M'(\Theta^*)(\Theta^t - \Theta^*) \qquad (1)$$

where $M'(\Theta^*) = \frac{\partial M}{\partial \Theta}|_{\Theta=\Theta^*}$. Since $M'(\Theta^*)$ is typically nonzero, a bound optimizer can essentially be seen as a linear iteration algorithm with a "convergence rate matrix" $M'(\Theta^*)$. Intuitively, $M'(\Theta^*)$ can be viewed as an operator that forms a contraction mapping around $\Theta^*$. In general, we would expect the Hessian $\frac{\partial^2 L(\Theta)}{\partial \Theta^2}|_{\Theta=\Theta^*}$ to be negative semidefinite, or negative definite, and thus the eigenvalues of $M'(\Theta^*)$ to all lie in $[0, 1]$ or $[0, 1)$ respectively [4]. Exceptions to the convergence of the bound optimizer to a local optimum of $L(\Theta)$ occur if $M'(\Theta^*)$ has eigenvalues whose magnitudes exceed unity.

Near a local optimum, this matrix is related to the curvature of the functional $G(\Theta, \Psi)$:

$$\lim_{\Theta^t \to \Theta^*} M'(\Theta^t) = -[\nabla_G^2(\Theta^*, \Psi^*)][\nabla_G^2(\Theta^*)]^{-1} \qquad (2)$$

where we define the mixed partials and Hessian as:

$$\nabla_G^2(\Theta^*, \Psi^*) \equiv \left[ \frac{\partial^2 G(\Theta, \Psi)}{\partial \Theta \partial \Psi^T} \Big| \begin{array}{l} \Theta = \Theta^* \\ \Psi = \Theta^* \end{array} \right] \qquad (3)$$

$$\nabla_G^2(\Theta^*) \equiv \left[ \frac{\partial^2 G(\Theta, \Psi)}{\partial \Theta \partial \Theta^T} \Big| \begin{array}{l} \Theta = \Theta^* \\ \Psi = \Theta^* \end{array} \right] \qquad (4)$$

We assume we can easily find $\arg\max_\Theta G(\Theta, \Psi)$, and thus $\nabla_G^2(\Theta^*)$ is negative definite (invertible).

( Proof sketch of eq (2): By performing Taylor series expansion of $\nabla G(\Theta^2, \Theta^1) = \frac{\partial G(\Theta, \Theta^1)}{\partial \Theta}|_{\Theta=\Theta^2}$ around $(\Theta^*, \Theta^*)$, we have: $\nabla G(\Theta^2, \Theta^1) = \nabla G(\Theta^*, \Theta^*) + (\Theta^2 - \Theta^*)^T \nabla_G^2(\Theta^*) + (\Theta^1 - \Theta^*)^T \nabla_G^2(\Theta^*, \Psi^*) + \ldots$. Substituting $\Theta^t$ for $\Theta^1$, and $M(\Theta^t)$ for $\Theta^2$ gives $0 = (M(\Theta^t) - \Theta^*)^T \nabla_G^2(\Theta^*) + (\Theta^t - \Theta^*)^T \nabla_G^2(\Theta^*, \Psi^*) + \ldots$. Assuming that higher order terms are negligible, in the limit, $\Theta^* = M(\Theta^*)$ and $0 = (\lim_{\Theta^t \to \Theta^*} M'(\Theta^t))\nabla_G^2(\Theta^*) + \nabla_G^2(\Theta^*, \Psi^*)$. )

What directions do bound optimizers move in parameter space? For most objective functions, the BO step $\Theta^{(t+1)} - \Theta^{(t)}$ in parameter space and true gradient vector $\nabla_L(\Theta^t) = \frac{\partial L(\Theta)}{\partial \Theta}|_{\Theta=\Theta^t}$ can be trivially related by a *transformation matrix* $P(\Theta^t)$, that changes at each iteration:

$$\Theta^{(t+1)} - \Theta^{(t)} = P(\Theta^t)\nabla_L(\Theta^t) \qquad (5)$$

Under certain conditions, this transformation matrix $P(\Theta^t)$ is guaranteed to be positive definite with respect to any gradient. In particular, **if C1:** $G(\Theta, \Theta^t)$ is well-defined, and differentiable everywhere in $\Theta$; **and C2:** for any fixed $\Theta^t \neq \Theta^{(t+1)}$, along any direction that passes through $\Theta^{t+1}$, $G(\Theta, \Theta^t)$ has only a single critical point in its first argument, located at the maximum $\Theta^{t+1}$; **then**

$$\nabla_L^T(\Theta^t) P(\Theta^t) \nabla_L(\Theta^t) > 0 \quad \forall \Theta^t \qquad (6)$$

The second condition may seem very strong, however, it is satisfied in many practical cases. For example, for the EM algorithm, it is satisfied whenever the M-step has a single unique solution (in particular, it holds for exponential family models due to concavity of $G(\Theta, \Theta^t)$); for GIS, NMF, CCCP, and many others, it is satisfied due to concavity of $G(\Theta, \Theta^t)$ (although C2 does not imply concavity).

(Proof sketch of eq (6): For $\nabla_G^T(\Theta^t)(\Theta^{(t+1)} - \Theta^t)$, we note that $\nabla_G^T(\Theta^t) = \frac{\partial G(\Theta, \Theta^t)}{\partial \Theta}|_{\Theta=\Theta^t}$ is the directional derivative of function $G(\Theta, \Theta^t)$ in the direction of $\Theta^{(t+1)} - \Theta^t$. C1 and C2 together imply that this quantity is positive, otherwise by the Mean Value Theorem (C1) $G(\Theta, \Theta^t)$ would have a critical point along some direction, located at a point other than $\Theta^{t+1}$ (C2). By using the identity $\nabla_L(\Theta^t) = \frac{\partial G(\Theta, \Theta^t)}{\partial \Theta}|_{\Theta=\Theta^t}$, we have $\nabla_L^T(\Theta^t) P(\Theta^t) \nabla_L(\Theta^t) = \nabla_G^T(\Theta^t)(\Theta^{(t+1)} - \Theta^t) > 0$. )

The important consequence of the above analysis is that when the bound function has a unique optimum wrt its first argument, BO has the appealing quality of always taking a step $\Theta^{(t+1)} - \Theta^t$ having positive projection onto the true gradient of the objective function $L(\Theta^t)$. This makes BO similar to a first order method operating on the gradient of a locally reshaped likelihood function.

For maximum likelihood learning of a mixture of Gaussians model using the EM-algorithm, this positive definite transformation matrix $P(\Theta^t)$ was first described by Xu and Jordan[11]. We have extended their results by deriving the explicit form of the transformation matrix for several other latent variables models such as Factor Analysis (FA), Probabilistic Principal Component Analysis (PPCA), mixture of PPCAs, mixture of FAs, and Hidden Markov Models [8]; we have also derived the general form of $P(\Theta^t)$ matrix for exponential family models in terms of natural parameters.

Here we further study the structure of the transformation matrix $P(\Theta^t)$ and relate it to the convergence rate matrix $M'$. Our main result is that when the derivative is small ($M'$ has small eigenvalues), the transformation matrix approaches the negative inverse Hessian and bound optimization behaves like a second-order Newton method. In particular, in the neighborhood of a local optimum $\Theta^*$:

$$\lim_{\Theta^t \to \Theta^*} P(\Theta^t) = \left[I - M'(\Theta^*)\right]\left[-S(\Theta^*)\right]^{-1} \qquad (7)$$

where $S(\Theta^*) = \frac{\partial^2 L(\Theta)}{\partial \Theta^2}|_{\Theta=\Theta^*}$ is the Hessian of the objective function. We assume that $P(\Theta)$ and $M(\Theta)$ are differentiable and that $[-S(\Theta^*)]^{-1}$ exists.

( Proof sketch of eq (7): Taking negative derivatives of (5) wrt $\Theta^t$ yields $I - M'(\Theta^t) = -P'(\Theta^t)\nabla_L(\Theta^t) - P(\Theta^t)S(\Theta^t)$ where $M'_{ij}(\Theta^t) = \partial\Theta_i^{t+1}/\partial\Theta_j^t$ is the input-output derivative matrix for the BO mapping and $P'(\Theta^t) = \frac{\partial P(\Theta^t)}{\partial \Theta}|_{\Theta=\Theta^t}$ is the tensor derivative of $P(\Theta^t)$ with respect to $\Theta^t$. In the limit, near a fixed point, the first term will vanish since the gradient is going to zero (assuming $P'(\Theta^t)$ does not become infinite); the equality (7) readily follows. )

This shows that the nature of the quasi-Newton behavior is controlled by the convergence matrix $M'(\Theta^*)$. When the matrix $M'$ has small eigenvalues, then near a local optimum bound optimization may exhibit quasi-Newton con-vergence behavior. This is also true in "plateau" regions where the gradient is very small even if they are not near a local optimum.

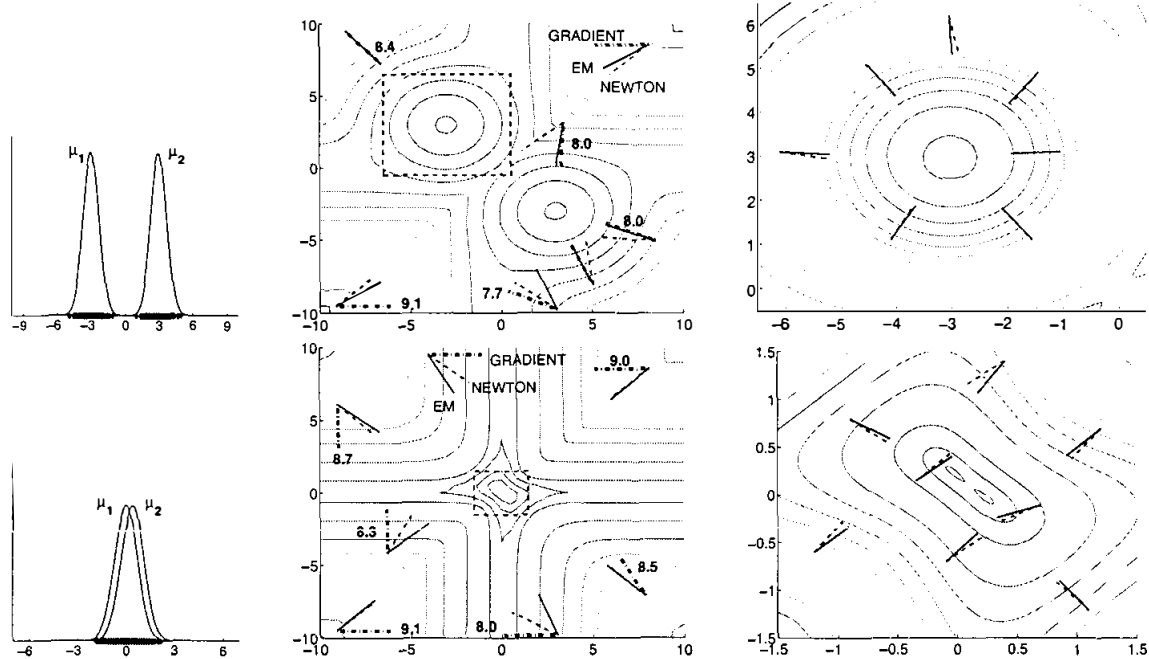

Figure 1: Contour plots of the likelihood function $L(\Theta)$ for MoG examples using well-separated (upper panels) and not-well-separated (lower panels) one-dimensional data sets. Axes correspond to the two means. The dashdot line shows the direction of the true gradient $\nabla_L(\Theta)$, the solid line shows the direction of $P(\Theta)\nabla_L(\Theta)$ and the dashed line shows the direction of $(-S)^{-1}\nabla_L(\Theta)$. Right panels are blowups of dashed regions on the left. The numbers indicate the log of the $l_2$ norm of $\nabla_L(\Theta)$. For the "well-separated" case, in the vicinity of $\Theta^*$, vectors $P(\Theta)\nabla_L(\Theta)$ and $(-S)^{-1}\nabla_L(\Theta)$ become identical.

We can examine the structure of this matrix and its eigenvalues, or the ratio of its two top eigenvalues. In particular, if the top eigenvalue of $M'(\Theta^*)$ tends to zero, then BO becomes a true Newton method, rescaling the gradient by exactly the negative inverse Hessian.

$$\Theta^{t+1} = \Theta^t - S(\Theta^t)^{-1}\nabla_L(\Theta^t) \qquad (8)$$

As the eigenvalues tend to unity, BO takes smaller and smaller stepsizes, giving poor, first-order convergence.

## 3 Common Bound Optimizers
### 3.1 Expectation-Maximization (EM)

We now consider a particular bound optimizer, the popular Expectation-Maximization (EM) algorithm, and derive specific cases of the results above for models which use EM to adjust their parameters. To begin, consider a probabilistic model of observed data $x$ which uses latent variables $y$. For any value of $\Psi$, it can be easily verified that the following difference of two terms is a lower bound on the likelihood:

$$G(\Theta, \Psi) = Q(\Theta, \Psi) - H(\Psi, \Psi) =$$
$$\int p(y|x, \Psi) \ln p(x, y|\Theta) dy - \int p(y|x, \Psi) \ln p(y|x, \Psi) dy$$

The log likelihood function can be written as:

$$L(\Theta) = \ln p(x|\Theta) = \int p(y|x, \Theta) \ln p(x|\Theta) dy$$
$$= G(\Theta, \Theta) \geq G(\Theta, \Psi) \quad \forall \Psi$$

By (2), we can establish:[1]

$$\nabla_G^2(\Theta^*) = \frac{\partial^2 Q(\Theta, \Theta^*)}{\partial \Theta^2}\Big|_{\Theta=\Theta^*}$$

$$\nabla_G^2(\Theta^*, \Psi^*) = -\frac{\partial^2 H(\Theta, \Theta^*)}{\partial \Theta^2}\Big|_{\Theta=\Theta^*}$$

and therefore we have an expression for $M'(\Theta^*)$:

$$\frac{\partial M(\Theta)}{\partial \Theta}\Big|_{\Theta=\Theta^*} = \left[\frac{\partial^2 H(\Theta, \Theta^*)}{\partial \Theta^2}\Big|_{\Theta=\Theta^*}\right]\left[\frac{\partial^2 Q(\Theta, \Theta^*)}{\partial \Theta^2}\Big|_{\Theta=\Theta^*}\right]^{-1}$$

This can be interpreted as the ratio of missing information to complete information near the local optimum [2, 5]. Notice that the curvature of the original bound function appears as one of the terms in the ratio. According to (7), in the neighborhood of a solution (for sufficiently large $t$):

$$P(\Theta^t) \approx \left[I - \left(\frac{\partial^2 H}{\partial \Theta^2}\right)\left(\frac{\partial^2 Q}{\partial \Theta^2}\right)^{-1}\Big|_{\Theta=\Theta^t}\right]\left[-S(\Theta^t)\right]^{-1}$$

The interpretation of this result is intuitive and well known: *When the missing information is small compared to the complete information, EM exhibits quasi-Newton behavior and enjoys fast, typically superlinear, convergence in the neighborhood of $\Theta^*$.* If the fraction of missing information approaches unity, the eigenvalues of the first term above approach zero and EM will exhibit extremely slow convergence. The above analysis gives a formal explanation (applicable to any latent variable model) of this behaviour.

Figure 1 illustrates these results in the case of fitting a mixture of Gaussians model to well-clustered and not-well-clustered data. Many other models also show this same effect; for example, when Hidden Markov Models or Aggregate Markov Models [9] are trained on very structured sequences, EM exhibits quasi-Newton behavior, in particular when the state transition matrix is sparse and the output distributions are almost deterministic at each state.

---
[1] For further details refer to [4]

### 3.2 Generalized Iterative Scaling (GIS)

In this section we consider the Generalized Iterative Scaling algorithm [1], widely used for parameter estimation in maximum entropy models. Its goal is to determine the parameters $\Theta^*$ of an exponential family distribution $p(x|\Theta) = \frac{1}{Z(\Theta)} \exp(\Theta^T F(x))$ such that certain generalized marginal constraints are preserved: $\sum_x p(x|\Theta^*) F(x) = \sum_x \bar{p}(x) F(x)$, where $Z(\Theta)$ is the normalizing factor, $\bar{p}(x)$ is a given empirical distribution and $F(x) = [f_1(x), ..., f_d(x), 1]^T$ is a given feature vector on the inputs. (We include the constant, or bias feature.) The GIS algorithm requires that $f_i(x) > 0 \ \forall i$ (but we will not require $\sum_i f_i(x) = 1$)[6]. The log-likelihood is:

$$L(\Theta) = \sum_x \bar{p}(x) \ln p(x|\Theta) = \sum_x \bar{p}(x)\Theta^T F(x) - \ln Z(\Theta)$$

We note that $\ln Z(\Theta) \leq Z(\Theta)/Z(\Psi) + \ln Z(\Psi) - 1$ for any $\Psi$, and $\exp \sum_i \Theta_i f_i(x) \leq \sum_i f_i(x) \exp \Theta_i + [1 - \sum_i f_i(x)]$, with $\sum_i f_i(x) \leq 1$. Defining $s = \max_x \sum_i f_i(x)$, we construct a lower bound:

$$L(\Theta) \geq \sum_x \bar{p}(x) \sum_i \Theta_i f_i(x) - \ln Z(\Psi) + \sum_i \frac{f_i(x)}{s} -$$
$$\sum_x p(x|\Psi) \sum_i \frac{f_i(x)}{s} \exp[s(\Theta_i - \Psi_i)] = G(\Theta, \Psi)$$

This lower bound has the useful property that its maximization is decoupled across the parameters $\Theta_i$. The GIS algorithm is then given by:

$$\Theta_i^{t+1} = \Theta_i^t + \frac{1}{s} \ln \frac{\sum_x \bar{p}(x) f_i(x)}{\sum_x p(x|\Theta^t) f_i(x)}$$

Define $\bar{F}(\Theta^*) \equiv \sum_x p(x|\Theta^*)F(x)$ to be the mean of the feature vectors, $D(\Theta^*) \equiv \text{diag}[\bar{F}(\Theta^*)]$ to be the corresponding diagonal matrix, and $\text{Cov}(\Theta^*)$ to be covariance of the feature vectors under model distribution $p(x|\Theta^*)$. We can compute second order statistics using (2):

$$\nabla^2_G(\Theta^*) = -s \, \text{diag}[\bar{F}(\Theta^*)] = -s\text{D}(\Theta^*)$$
$$\nabla^2_G(\Theta^*, \Psi^*) = s \, \text{diag}[\bar{F}(\Theta^*)] -$$
$$\left[\sum_x p(x|\Theta^*)F(x)F(x)^T - [\bar{F}(\Theta^*)][\bar{F}(\Theta^*)]^T\right]$$
$$= s\text{D}(\Theta^*) - \text{Cov}(\Theta^*)$$

According to (7), in the neighborhood of a solution (for sufficiently large $t$), the step GIS takes in parameter space and true gradient are related by the matrix:

$$P(\Theta^t) \approx \left[\frac{1}{s}\text{Cov}(\Theta^t)\text{D}(\Theta^t)^{-1}\right]\left[-S(\Theta^t)\right]^{-1}$$

Due to the concavity of $G(\Theta, \Psi')$ for any fixed $\Psi'$, the step a GIS algorithm takes in parameter space always has positive projection onto the true gradient of the objective function. The convergence rate matrix $M'(\Theta^*)$ is of the form:

$$\frac{\partial M(\Theta)}{\partial \Theta}\Big|_{\Theta=\Theta^*} = I - \frac{1}{s}\text{Cov}(\Theta^*)\text{D}(\Theta^*)^{-1} \quad (9)$$

and depends on the covariance and the mean of the feature vectors. We can interpret this result as follows: *when feature vectors become less correlated and closer to the origin, GIS exhibits faster convergence in the neighborhood of* $\Theta^*$. If features are highly dependent, then GIS will exhibit extremely slow convergence.

### 3.3 Non-Negative Matrix Factorization (NMF)

Given a non-negative matrix V, the NMF algorithm[3] tries to find matrices W and H, such that $V \approx WH$. Posed as an optimization problem, we are interested in minimizing a divergence $L(W, H) = D(V||WH)$, subject to $(W, H) \geq 0$ elementwise:

$$L(W, H) = \sum_{ij} \left(V_{ij} \ln \frac{V_{ij}}{(WH)_{ij}} - V_{ij} + (WH)_{ij}\right)$$

We use $-\ln \sum_c W_{ic}H_{cj} \leq -\sum_c \alpha_{ij}(c,c) \ln \frac{W_{ic}H_{cj}}{\alpha_{ij}(c,c)}$ where $\alpha_{ij}(a,b) = W_{ia}^t H_{bj}^t / \sum_r W_{ir}^t H_{rj}^t$, so that $\alpha_{ij}(c,c)$ sum to one. Defining $\Theta = (W, H)$ and $\Psi = (W^t, H^t)$, we can construct the upper bound on the cost function:

$$L(\Theta) \leq \sum_{ij} V_{ij} \ln V_{ij} - V_{ij} + \sum_{ijc} W_{ic}H_{cj} - \quad (10)$$
$$\sum_{ijc} V_{ij}\alpha_{ij}(c,c)\left[\ln \frac{W_{ic}H_{cj}}{\alpha_{ij}(c,c)}\right] = G(\Theta, \Psi)$$

One can now compute second order statistics using (2). In the appendix we derive the explicit form of the convergence rate matrix $M'$. We also note that the convergence matrix of NMF much resembles the convergence matrix of GIS, since both algorithms make use of the bound that comes from Jensen's inequality.

### 3.4 Concave-Convex Procedure (CCCP)

A CCCP [12] optimizer seeks to minimize an energy function $E(\Theta)$, which can be decomposed into a convex $E_{vex}(\Theta)$ and a concave $E_{cave}(\Theta)$ function:

$$E(\Theta) = E_{vex}(\Theta) + E_{cave}(\Theta) \quad (11)$$

CCCP algorithm is given by:

$$\nabla E_{vex}(\Theta^{t+1}) = -\nabla E_{cave}(\Theta^t)$$

It is easy to see that CCCP belongs to the class of bound optimization algorithms, and therefore can be analyzed as a first order iterative algorithm. Its bound function is:

$$E(\Theta) \leq E_{vex}(\Theta) + E_{cave}(\Psi) +$$
$$(\Theta - \Psi)^T \nabla E_{cave}(\Psi) = G(\Theta, \Psi)$$

Employing (2), we have:

$$\nabla^2_G(\Theta^*) = \frac{\partial^2 E_{vex}(\Theta)}{\partial \Theta \partial \Theta^T}\Big|_{\Theta=\Theta^*}$$
$$\nabla^2_G(\Theta^*, \Psi^*) = \frac{\partial^2 E_{cave}(\Psi)}{\partial \Psi \partial \Psi^T}\Big|_{\Psi=\Theta^*}$$

The convergence rate matrix is given by:

$$M'(\Theta^*) = -\left[\frac{\partial^2 E_{cave}(\Psi)}{\partial \Psi \partial \Psi^T}\Big|_{\Psi=\Theta^*}\right]\left[\frac{\partial^2 E_{vex}(\Theta)}{\partial \Theta \partial \Theta^T}\Big|_{\Theta=\Theta^*}\right]^{-1}$$

which can be interpreted as a ratio of concave curvature to convex curvature. According to (7) in the neighborhood of a solution (for sufficiently large $t$) the gradient and step are related by: $P(\Theta^t) \approx$

$$\left[I - \left(\frac{\partial^2 E_{cave}}{\partial \Theta^2}\right)\left(\frac{\partial^2 E_{vex}}{\partial \Theta^2}\right)^{-1}|_{\Theta=\Theta^t}\right]\left[-S(\Theta^t)\right]^{-1}$$

Of course, the step CCCP takes in parameter space has positive projection onto the true gradient of the original energy function $E(\Theta)$.

The above view of CCCP has an interesting interpretation: *If the concave energy function has small curvature compared to the convex energy term in the neighborhood of $\Theta^*$, CCCP will exhibit a quasi-Newton behavior and will possess fast, typically superlinear convergence.* As the fraction of concave-convex curvature approaches one, CCCP will exhibit extremely slow, first order convergence behavior. Figure 4 illustrates exactly such an example.

## 4 Improving Convergence Rates

The above analysis helped to answer the question: when and why will bound optimizers converge slowly? They can also help to answer the more practical question: what can we do to speed up convergence?

In the case of EM, it is possible to estimate the key quantity controlling convergence (fraction of missing information) and switch to direct (gradient-based) optimization when we predict slow behavior of EM. We have experimented with such a "hybrid" approach with some success[7]. For other bound optimizers, similar hybrid algorithms are possible.

But there is another, intriguing approach to improving convergence speed: modify the original input to the algorithms based on our analysis of convergence rates. In the case of GIS this involves transforming features, in the case of NMF, this requires scaling and translating data vectors, and for CCCP this comes down to designing different convex-concave decompositions of the objective. These input modifications do not change the final results of the algorithms; they only change the convergence properties.

Beginning with GIS, we can show that *translating feature vectors to bring them closer to the origin and decorrelating (whitening) them* both speed up convergence. (Homogeneously rescaling all features by a single constant does not affect convergence.) In particular, the optimal translation of features is given by $F_{new}(x) = F(x) - V$ with $V_i = \min_x f_i(x) \, \forall i$, and the optimal linear transformation $AF_{new}$ is that which makes $A\text{Cov}(\Theta^*)A^T$ equal to identity matrix, taking into account the bias term, or a feature that is a constant. (We provide sketch proofs of both results in the appendix.) Of course, the covariance in the second condition cannot be evaluated until the optimal parameters are known, but it can be approximated by using the sample covariance of features on the training set.

For NMF, similar to GIS, we can show that *translating data vectors to bring them closer to the origin* speeds up convergence, whereas homogeneously rescaling all data by a single constant does not affect convergence.

For CCCP, it is well-known that any energy function with bounded curvature has many convex-concave decompositions but no clear principle for finding a good one has been known. Our analysis provides guidance in this regard: we should *minimize the ratio of curvatures between the convex and concave parts of the energy.*

In the next section we illustrate that appropriate preprocessing of the input to these various bound optimization algorithms *does* result in a much faster rate of convergence.

## 5 Experimental Results

We now present empirical results to support the validity of our analysis for several bound optimization algorithms. We first apply EM to learning the parameters of two latent variable models: Mixtures of Gaussians (MoG) and Hidden Markov Models (HMM). We then analyze and apply Iterative Scaling (IS) to a logistic regression model. Next, we show the effect of data translation on the convergence properties of NMF. Finally, we finish by describing and analyzing the effect of various energy function decompositions on the convergence behavior of the CCCP algorithm. Though not shown, we confirmed that the convergence results presented below do not vary significantly for different random initial starting points in the parameter space.

First, consider a mixture of Gaussians (MoG) model. In this model the proportion of missing information corresponds to how "well" or "not-well" the data is separated into distinct clusters. We therefore considered two types of data sets, a "well-separated" case and a "not-well-separated" case in which the data overlaps in one contiguous region. As predicted by our analysis, in the "well-separated" case, in the vicinity of the local optimum $\Theta^*$ the directions of the vectors $P(\Theta)\nabla_L(\Theta)$ and $(-S)^{-1}\nabla_L(\Theta)$ become identical (fig. 1), showing that EM will have quasi-Newton convergence behavior. In "not-well-separated" case, due to the large proportion of missing information, these directions are significantly different and EM possesses poor, first-order convergence behavior.

We also applied the MoG model to cluster a set of 50,000 $8 \times 8$ greyscale pixel image patches.[2] Figure 2 displays the convergence behavior of EM for M=5 and M=50 mixture components. The experimental results reveal, that with fewer mixture components, EM converges quickly to a local optimum, since the components generally model the data with fairly distinct, non-contiguous clusters. As the number of mixtures components increases, clusters overlap in contiguous regions, creating a relatively high proportion of missing information. In this case the convergence of EM slows by several orders of magnitude.

We then applied EM to training Hidden Markov Models (HMMs). Missing information in this model is high when the observed data do not well determine the underlying state sequence (given the parameters). We therefore generated two synthetic data sets from a 5-state HMM, with an alphabet size of 5 characters. The first data set ("aliased" sequences) was generated from a HMM where output parameters were set to uniform values plus some small noise $\epsilon \sim \mathcal{N}(0, 01I)$. The second data set ("structured sequences") was generated from a HMM with sparse transition and output matrices. Figure 2 shows that for the very structured data, EM performs well and exhibits second order convergence in the vicinity of the local optimum. For the ambiguous or aliased data, EM posses extremely slow, first-order convergence behavior.

---

[2]The data set used was the **imlog** data set publicly available at ftp://hlab.phys.rug.nl/pub/samples/imlog

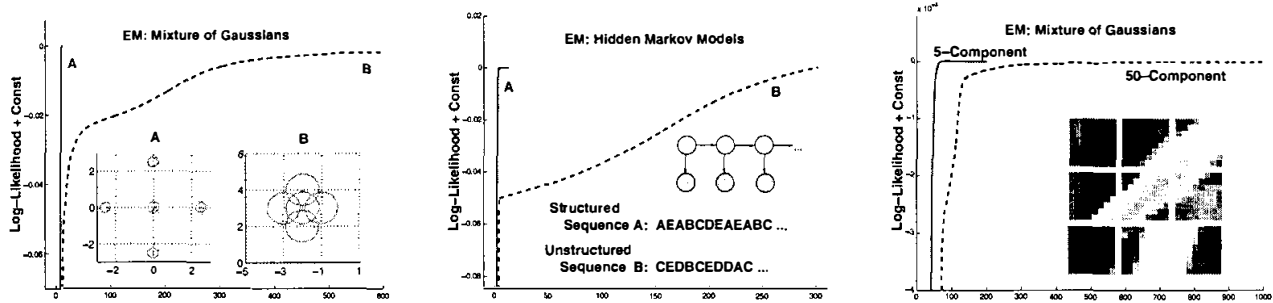

Figure 2: Learning curves of EM algorithm for two models: MoG and HMM. Different data sets are shown on the same plots for convenience. The iteration number is shown on the horizontal axis, and log-likelihood is shown on the vertical axis with the zero-level likelihood corresponding to the converging point of the EM algorithm. For "well-separated" and "structured" data (A), EM possesses quasi-Newton convergence behavior. EM in this case converges in 10-15 iterations with stopping criterion: $[L(\Theta^{t+1}) - L(\Theta^t)]/\text{abs}(L(\Theta^{t+1})) < 10^{-15}$. For "overlapping", "aliased" data (B), EM posses poor, first-order convergence. Right panel displays convergence behavior of EM by fitting 5 component as opposed to 50 component MoG model on the same data set of gray image patches.

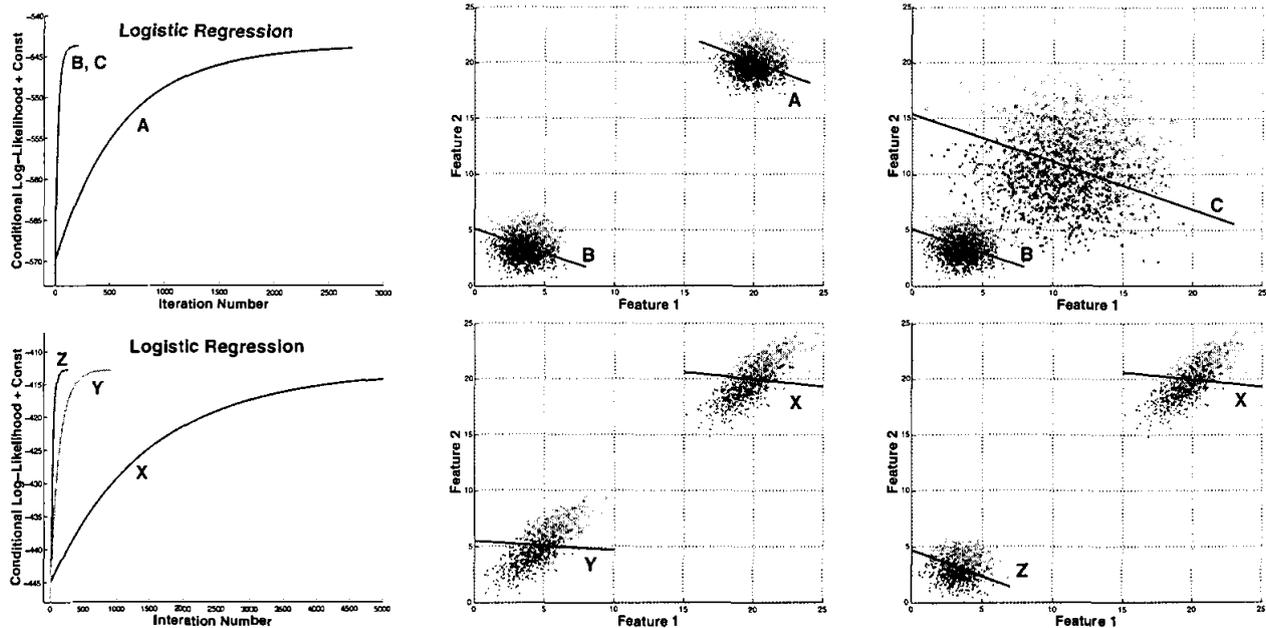

Figure 3: Learning curves (left panels) of Iterative Scaling algorithm for logistic regression model, showing the effect that translation and whitening of the feature vectors have on the IS convergence behavior, with letters corresponding to the respective data sets. Top panels show an experiment with 2,000 2-dimensional feature vectors drawn from standard normal, bottom panels display an identical experiment with 2,000 feature vectors drawn from normal with oriented covariance. Top, right panel shows that scaling feature vectors by constant does not affect the convergence of IS.

This analysis may also shed light on why hard-clustering algorithms such as k-means and Viterbi style E-steps for HMMs appear to have faster convergence than their softer cousins: they suppress the missing information.

To confirm our analysis of GIS, we applied iterative scaling algorithm to a simple 2-class logistic regression model: $p(y = \pm 1|x, w) = 1/(1 + \exp(-yw^T x))$ following [6]. In our first experiment, $N$ feature vectors of dimensionality $d$ were drawn from normal: $x \sim \mathcal{N}(0, 2I_d)$, with the true parameter vector $w^*$ being randomly chosen on the surface of the $d$-dimensional sphere with radius $\sqrt{2}$. To make features positive, the data set was modified by adding 20 to all feature values. Figure 3 shows that for $N = 2000$ and $d = 2$, naive IS, that runs on the original unpreprocessed features, takes over 2500 iterations to converge. When feature vectors are translated closer to the origin, IS converges to exactly the same maximum likelihood solution, but beats naive IS by a factor of almost twelve.

Our second experiment was similar, but feature vectors of dimensionality $d$ were drawn from a Gaussian with ori-ented covariance. Figure 3 shows that for N=2,000 and d=2, translating features improves the convergence of IS by a factor of over 4, whereas translating and whitening feature vectors results in speedup by factor of twenty. Similar results are obtained if dimensionality of data is increased.

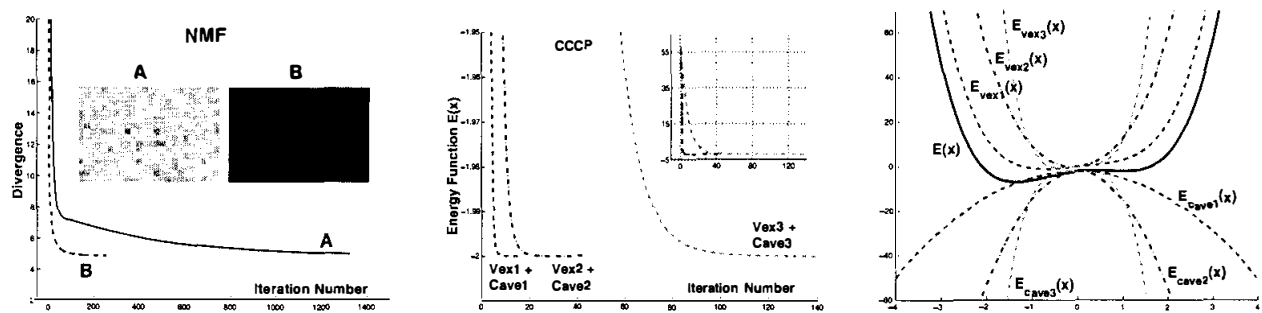

Figure 4: Learning curves of NMF and CCCP algorithms. For NMF, we show the effect that data translation has on the convergence behavior of NMF (in our case black pixels correspond to 0, white to 30). Applying CCCP to minimize a simple energy function $E(x) = x^4 - 3x^2 + 2x - 2$, we display the effect that different energy decompositions (left panel) have on CCCP convergence.

Next, we experimented with the NMF algorithm. Data vectors were drawn from standard normal: $x \sim \mathcal{N}(0, I_{16})$. To make features positive, the data set was modified by adding 20 to all data values, forming non-negative matrix $V$. We then applied NMF to perform non-negative factorization: $V \approx WH$. Figure 4 reveals that naive NMF, that runs on the original unpreprocessed data (data set A), takes over 1,300 iterations to converge. Once data vectors are translated closer to the origin (data set B), NMF converges to exactly the same value of the cost function in about 230 iterations, outperforming naive NMF by a factor of over five.

Finally, we experimented with the CCCP algorithm. We considered a simple energy function $E(x)=x^4-3x^2+2x-2$, which has many decompositions (fig.4). A decomposition which minimizes the ratio of concave-convex curvature is: $E_{cave1}(x)=-3x^2-2$ and $E_{vex1}(x)=x^4+2x$. Other decompositions: $E_{cave2}(x)=-13x^2-2$ and $E_{vex2}(x)=x^4+10x^2+2x$; $E_{cave3}(x)=-9x^4-3x^2-2$ and $E_{vex3}(x)=10x^4+2x$; clearly increase the proportion of concave-convex curvature. In our experiment, all runs of CCCP were started from the same initial point in the parameter space. Figure 4 reveals that as the proportion of the local concave-convex curvature increases, the convergence rate of CCCP significantly slows down, by several orders of magnitude.

## 6 Discussion

In this paper we have analyzed a large class of bound optimization algorithms and their relationship to direct optimization algorithms such as gradient-based methods. We determined conditions under which BO algorithms exhibit local-gradient and fast quasi-Newton convergence behaviors. Based on this analysis and interpretation, we have also provided some recommendations for how the input to these algorithms can be preprocessed to yield faster convergence. Currently, using derivation of an explicit form of the convergence rate matrix, we are also working on identifying analytic conditions under which CCCP possesses fast or extremely slow convergence in minimizing Bethe and Kikuchi free energies in approximate inference problems. Similar analysis can be applied to other bound optimization algorithms; for example Sha et. al. [10] recently introduced a multiplicative algorithm for training SVMs and provided a convergence analysis based on margins.

Our analysis and experiments show that in the regime where the convergence rate matrix has large eigenvalues, a bound optimizer is likely to perform poorly. Slow convergence is expected when missing information is high while learning with EM; when feature vectors are highly dependent while estimating parameters with GIS or NMF; or when the ratio of concave-convex curvature is large when minimizing energy function with CCCP. In these cases, one can either attempt to modify the basic BO algorithms to accelerate them, or instead employ direct optimization algorithms such as conjugate-gradient which are likely to have far superior performance. Alongside our analysis we have also presented a third alternative: inputs to standard BO algorithms can sometimes be preprocessed to speed convergence.

### Acknowledgments
Funded in part by the IRIS project, Precarn Canada.

# A  Appendix

**Claim 1:** Translating feature vectors closer to the origin speeds up convergence of GIS. The optimal translation of features is given by $F_{new}(x) = F(x) - V$ $\forall x$ with a vector $V$ containing elements $V_i = \min_x f_i(x)$ $\forall i$.

Proof sketch: Consider setting $F_{new}(x) = F(x) - V$ $\forall x$ as above. We have

$$M'_{new}(\Theta^*) = I - \frac{1}{s_{new}} \text{Cov}(\Theta^*) D_{new}(\Theta^*)^{-1} \quad (12)$$

with $D_{new}(\Theta^*) = D(\Theta^*) - \text{diag}(V)$ (see eq. 9), and $s_{new} = s - \sum_i V_i$. Let us denote $Q(\Theta^*) \equiv \text{Cov}(\Theta^*) D(\Theta^*)^{-1}$, $Q_{new}(\Theta^*) \equiv \text{Cov}(\Theta^*) D_{new}(\Theta^*)^{-1}$, and $\lambda_{max}(A) \equiv$ the largest eigenvalue of $A$. We can now show that this translation forces the top eigenvalue of $M'(\Theta^*)$ to decrease: $\lambda_{max}(M'_{new}(\Theta^*)) \leq \lambda_{max}(M'(\Theta^*))$, where we derived (9): $M'(\Theta^*) = I - \frac{1}{s} Q(\Theta^*)$. Note that: $\lambda_{max}(M'_{new}(\Theta^*)) = 1 - \lambda_{min}\left(\frac{1}{s_{new}} Q_{new}(\Theta^*)\right)$. Hence, our task reduces to showing:

$$\lambda_{min}\left(\frac{1}{s_{new}} Q_{new}(\Theta^*)\right) \geq \lambda_{min}\left(\frac{1}{s} Q(\Theta^*)\right)$$

$$\Rightarrow \lambda_{max}\left(s_{new} Q_{new}^{-1}(\Theta^*)\right) \leq \lambda_{max}\left(s Q^{-1}(\Theta^*)\right) \quad (13)$$

Taking into account that $s_{new} \leq s$, the above inequality is obvious by examining:

$$\lambda_{max}\left(s_{new} Q_{new}^{-1}(\Theta^*)\right) =$$
$$s_{new} \lambda_{max}\left([D(\Theta^*) - \text{diag}(V)]\text{Cov}^{-1}(\Theta^*)\right) \leq$$
$$s \lambda_{max}\left(D(\Theta^*)\text{Cov}^{-1}(\Theta^*)\right) = s \lambda_{max}\left(Q^{-1}(\Theta^*)\right)$$

It is now clear that the optimal translation of features is given by $F_{new}(x) = F(x) - V$ $\forall x$ with $V_i = \min_x f_i(x)$ $\forall i$.

**Claim 2:** Decorrelating (whitening) feature vectors speeds up convergence of GIS In particular, the optimal linear transformation $F_{new}(x) = AF(x)$ is that which makes $A\text{Cov}(\Theta^*)A^T$ equal to identity matrix.

Proof sketch: Consider spectral decomposition: $\text{Cov}(\Theta^*) = WHW^T$, with $H$ being the diagonal matrix of the eigenvalues, and $W$ being the orthogonal matrix of the corresponding eigenvectors. Let $A = WH^{-1/2}W^T$. The linear transformation becomes $F_{new}(x) = AF(x)^3$, in which case $A\text{Cov}(\Theta^*)A^T = I$.

$$M'_{new}(\Theta^*) = I - \frac{1}{s_{new}} A\text{Cov}(\Theta^*) A^T D_{new}(\Theta^*)^{-1}$$
$$= I - \frac{1}{s_{new}} D_{new}(\Theta^*)^{-1} \quad (14)$$

with $s_{new} = \max_x \sum_i [AF(x)]_i$, and $D_{new}(\Theta^*) = \text{diag}[A \sum_x p(x|\Theta^*) F(x)] = \text{diag}[A\bar{F}(\Theta^*)]$. We now show that, in general, $\lambda_{max}(M'_{new}(\Theta^*)) \leq \lambda_{max}(M'(\Theta^*))$. This task reduces to showing (see eq (13)): $\lambda_{max}(s_{new} D_{new}(\Theta^*)) \leq \lambda_{max}(sQ^{-1}(\Theta^*))$. First note that:

$$\lambda_{max}(sQ^{-1}(\Theta^*)) = s\lambda_{max}(D(\Theta^*)\text{Cov}^{-1}(\Theta^*))$$
$$= s\lambda_{max}(D(\Theta^*)AA^T) \quad (15)$$

On the other side:

$$s_{new}\lambda_{max}(D_{new}(\Theta^*)) = s_{new} \| A\bar{F}(\Theta^*) \|_\infty \leq$$
$$s_{new} \| D(\Theta^*)A \|_\infty \quad (16)$$

It can also be shown that $s_{new} \leq s\lambda_{max}(A) = s \| A \|_2$. By using above facts, slightly more relaxed bound holds:

$$\| D(\Theta^*)A \cdot s_{new} \|_2 \leq \| D(\Theta^*)A \cdot sA \|_2 \quad (17)$$

Therefore in general, "whitening" feature vectors, pushes down the top eigenvalue of the convergence rate matrix, which according to our analysis, results in its faster rate of convergence.

**Non-Negative Matrix Factorization:** We use a bound on the objective function (10) to derive the explicit form of the convergence rate matrix $M'$. Defining $\Theta = (W, H)$ and $\Psi = (W^t, H^t)$, we employ (2):

$$\frac{\nabla_G^2(\Theta^*)}{\partial W_{ic}^* \partial W_{kp}^*} = \delta_{ik}\delta_{cp} \sum_j \frac{V_{ij}}{\bar{V}_{ij}^*} \frac{H_{cj}^*}{W_{ic}^*} \quad \frac{\nabla_G^2(\Theta^*)}{\partial W_{ic}^* \partial H_{pl}^*} = \delta_{cp}$$

$$\frac{\nabla_G^2(\Theta^*)}{\partial H_{cj}^* \partial H_{pl}^*} = \delta_{cp}\delta_{jl} \sum_i \frac{V_{ij}}{\bar{V}_{ij}^*} \frac{W_{ic}^*}{H_{cj}^*} \quad \frac{\nabla_G^2(\Theta^*)}{\partial H_{cj}^* \partial W_{kp}^*} = \delta_{cp}$$

$$\frac{\nabla_G^2(\Theta^*, \Psi^*)}{\partial W_{ic}^* \partial W_{kp}^*} = -\left[\delta_{ik}\delta_{cp}\sum_j \frac{V_{ij}}{\bar{V}_{ij}^*}\frac{H_{cj}^*}{W_{ic}^*} - \delta_{ik}\sum_j \frac{V_{ij}}{\bar{V}_{ij}^*}\frac{H_{cj}^*}{W_{ic}^*}\alpha_{ij}(c,p)\right]$$

$$\frac{\nabla_G^2(\Theta^*, \Psi^*)}{\partial H_{cj}^* \partial H_{pl}^*} = -\left[\delta_{jl}\delta_{cp}\sum_i \frac{V_{ij}}{\bar{V}_{ij}^*}\frac{W_{ic}^*}{H_{cj}^*} - \delta_{jl}\sum_i \frac{V_{ij}}{\bar{V}_{ij}^*}\frac{W_{ic}^*}{H_{cj}^*}\alpha_{ij}(c,p)\right]$$

$$\frac{\nabla_G^2(\Theta^*, \Psi^*)}{\partial W_{ic}^* \partial H_{pl}^*} = -\frac{V_{ij}}{\bar{V}_{ij}^*}(\delta_{cp} - \alpha_{il}(c,p))$$

$$\frac{\nabla_G^2(\Theta^*, \Psi^*)}{\partial H_{cj}^* \partial W_{kp}^*} = -\frac{V_{kj}}{\bar{V}_{kj}^*}(\delta_{cp} - \alpha_{kj}(c,p))$$

where we define $\bar{V}_{ij}^* = \sum_c W_{ic}^* H_{cj}^*$, and $\delta_{ij} = 1$ if $i = j$; 0 – otherwise. The convergence rate matrix $M'$ will be of the form:

$$\frac{\partial M(\Theta)}{\partial \Theta}\bigg|_{\Theta=\Theta^*} = -\left[\nabla_G^2(\Theta^*, \Psi^*)\right]\left[\nabla_G^2(\Theta^*)\right]^{-1}$$

---

[3]Here we are assuming that the new feature vector AF(x) has only positive entries. If AF(x) has negative entries it might be necessary to decorrelate and add a translation, which trades off the advantage of Claim 1 and Claim 2.